\documentclass{article}

\usepackage[main, final]{neurips_2026}

\usepackage[utf8]{inputenc}
\usepackage[T1]{fontenc}
\usepackage{hyperref}
\usepackage{url}
\usepackage{booktabs}
\usepackage{amsfonts}
\usepackage{amsmath}
\usepackage{nicefrac}
\usepackage{microtype}
\usepackage{xcolor}
\usepackage{graphicx}
\usepackage{subcaption}
\usepackage{multirow}

\newcommand{\pp}{\,\textrm{pp}}
\newcommand{\dtheta}{d_{\theta}}
\newcommand{\dthetal}{d_{\theta,l}}
\newcommand{\sigmas}{\sigma_{99}}
\newcommand{\pca}{\mbox{\textsc{Pca}}}
\newcommand{\uni}{\mbox{\textsc{Uniform}}}
\newcommand{\inv}{\mbox{\textsc{Inverted}}}
\newcommand{\ev}{\mbox{\textsc{Pca-Ev}}}
\newcommand{\slim}{\mbox{\textsc{SlimLLM}}}

\title{Measuring Layer-wise Intrinsic Dimensionality\\
of FFNs in LLMs via PCA}

\author{Daehee Kim\\
NAVER Cloud\\
\texttt{daehee.k@navercorp.com}}

\begin{document}
\maketitle

\begin{abstract}
Modern large language models {(LLMs)} keep the feed-forward network (FFN) intermediate width identical across every layer, 
a convention inherited from the original Transformer. 
Yet mechanistic work suggests layers at different depths do qualitatively different work, raising a natural question: 
are their capacity demands also non-uniform, and if so, can a fixed parameter budget be allocated more wisely than uniformly? 
To probe this, we apply principal component analysis (PCA) to the input of each FFN down-projection and measure, 
for every layer, its intrinsic dimensionality as the smallest number of principal components needed to reconstruct the activation up to a fixed cosine similarity. 
Scanning 58 open-weight LLMs spanning 14 families and 70M to 72B parameters, 
we find a recurring middle-heavy profile in which middle layers carry higher intrinsic dimensionality than edge layers; 
the profile is robust to the calibration corpus, even across English and Chinese, 
and is largely preserved through post-training fine-tuning. As a direct test that this signal is structurally meaningful, 
we use it to guide width pruning in a knowledge-distilled student under a matched parameter budget: 
{the PCA-guided allocation outperforms the uniform-width convention by $+2.3$ points on Qwen3-8B and $+2.5$ on Qwen2.5-3B,
while lower-dispersion teachers show smaller gaps ($+0.9$ on Gemma-3-4B, $+0.5$ on LLaMA-3-8B), and a budget-matched inverted control does worst on all four teachers. 
Across these four teachers,} the size of the gap {tracks} how unevenly the teacher's intrinsic dimensionality is distributed across layers 
(a single-number summary we call $\sigmas$), and the advantage persists through downstream supervised fine-tuning.
{Code is available at \url{https://github.com/naver-ai/lidf}.}
\end{abstract}

\section{Introduction}
\label{sec:intro}

Every mainstream LLM~\citep{touvron2023llama,yang2024qwen2,jiang2023mistral} applies the same FFN intermediate dimension to every transformer block, a convention inherited from the original Transformer~\citep{vaswani2017attention}.
This uniformity is computationally convenient but embeds a strong implicit assumption: that every layer performs representational work of the same complexity.
A growing body of mechanistic work~\citep{geva2021transformer,meng2022rome,dai2022knowledge,geva2022ffn} suggests this assumption does not hold: layers at different depths process qualitatively different information.
Yet whether the per-layer capacity demand is also non-uniform, and whether deviating from uniform width yields measurable benefits, has not been directly characterized at scale.

We explore this gap by introducing a simple, data-agnostic measurement protocol for layer-wise {intrinsic dimensionality} and by validating its predictions in a controlled width-pruning experiment.
Specifically, we hook the input to each FFN's down-projection, the last intermediate activation of the layer in both gated (SwiGLU)~\citep{shazeer2020glu} and non-gated (Gaussian Error Linear Unit, GELU)~\citep{hendrycks2016gelu} FFNs.
We apply PCA~\citep{jolliffe2002pca} to it and define $\dtheta$ as the smallest rank at which the cosine similarity between the activation and its top-$d$ PCA reconstruction reaches a threshold $\theta$.
The result is a per-layer integer, comparable across layers and architectures, that requires only a single forward pass on a small unlabeled corpus.

Three observations emerge from a 58-model scan covering 14 families at 70M--72B parameters:
(i) the per-layer $\dtheta$ profile is recurringly middle-heavy: most models exhibit higher {$\dtheta$} at intermediate depth and lower {$\dtheta$} at the edges;
(ii) the profile is largely robust to the calibration corpus, including across English and Chinese inputs (per-layer Pearson $r \ge 0.94$ on Qwen3-8B~\citep{yang2025qwen3});
(iii) the layer-wise dispersion $\sigmas := \mathrm{std}_l(\dthetal / D_l)$ at the default threshold $\theta = 0.99$, a single-number summary of how unevenly intrinsic dimensionality is distributed across layers, is a model-internal property that is approximately preserved through post-training {such as supervised fine-tuning (SFT)}.

{As a functional test of the measurement, we use $\dtheta$} directly as a width-allocation rule for a {student trained by knowledge distillation (KD)}~\citep{hinton2015distilling}.
Three students with budget-matched FFN allocations---\pca{}-guided (each layer's width set to $\dthetal$), \uni{} (flat width, current convention), and \inv{} (point-wise mirror around uniform, an adversarial control)---are distilled from the same teacher with identical optimizer, data, and learning-rate schedule.
{On the two high-$\sigmas$ teachers (Qwen2.5-3B~\citep{qwen2024qwen25} and Qwen3-8B), \pca{} outperforms \uni{} by $+2.3$ to $+2.5$ percentage points (pp) and \inv{} by $+4.6$ to $+5.5$\pp{} on a five-task average; on the lower-$\sigmas$ Gemma-3-4B~\citep{team2025gemma3} and LLaMA-3-8B~\citep{grattafiori2024llama3} the gaps against \uni{} are smaller ($+0.5$ to $+0.9$\pp).}
The advantage {tracks layer-wise dispersion ($\sigmas$) across the four teachers} and is {negligible at 1.5B and below}; a downstream SFT round preserves the ordering, with $\pca{} - \uni{}$ at $+11.0$\pp{} on ARC-Challenge {(ARC-C)}~\citep{clark2018arc}, $+4.9$\pp{} on MMLU~\citep{hendrycks2021mmlu}, and $+3.0$\pp{} on GSM8K~\citep{cobbe2021gsm8k}.

\paragraph{Contributions.}
\begin{enumerate}
  \item \textbf{Measurement protocol.}
    We define a layer-wise intrinsic dimensionality $\dtheta$ for FFN down-projection input activations via PCA combined with a cosine-similarity threshold (\S\ref{sec:method}).
    The protocol is data-agnostic and label-free.
  \item \textbf{Empirical characterization across 58 LLMs.}
    We document a recurring middle-heavy profile, calibration-data robustness 
    (Pearson $r \ge 0.94$ across all pairs of 5 English corpora and Chinese SkyPile~\citep{wei2023skywork}), 
    a pretraining-trajectory growth phase {in 20 of 21 trajectories across six families}, and 
    post-training stability of the layer-wise dispersion (\S\ref{sec:what-d-measures}).
  \item {\textbf{Functional validation: controlled KD.}
    PCA-guided width allocation outperforms a budget-matched uniform allocation on four teachers at moderate-to-large scale; {the inverted control falls below the uniform allocation}; the advantage persists through SFT (\S\ref{sec:kd-validation}).}
\end{enumerate}

\section{Measurement Protocol}
\label{sec:method}

\paragraph{Where to measure.}
For a transformer layer $l$, the intermediate activation just before the down-projection is
\begin{equation}
  h_l \;=\;
  \begin{cases}
    \phi(W_g^{(l)} x_l) \;\odot\; (W_u^{(l)} x_l) & \text{(gated FFN)} \\
    \phi(W_u^{(l)} x_l) & \text{(non-gated FFN)}
  \end{cases}
  \;\in\; \mathbb{R}^{D_l},
  \label{eq:gated-ffn}
\end{equation}
where $x_l$ is the residual-stream input, $\phi$ is the layer's activation function (e.g., SiLU~\citep{elfwing2018silu} for SwiGLU, GELU for GeGLU~\citep{shazeer2020glu} or non-gated GELU FFNs), $\odot$ is elementwise product, and $D_l$ is the FFN intermediate dimension.
We hook $h_l$ at \texttt{model.layers[l].mlp.down\_proj.input} and treat $\{h_l\}$ as the object whose {intrinsic dimensionality} we measure.

\paragraph{Calibration set and PCA.}
We collect $N \approx 100\text{K}$ activation tokens per layer (50 sequences of 2{,}048 tokens) from SlimPajama~\citep{cerebras2023slimpajama}.
{Writing $\bar h_l$ for the per-layer activation mean, we compute the eigendecomposition of the centered covariance $\tilde X_l^{\!\top} \tilde X_l / (N-1) \in \mathbb{R}^{D_l \times D_l}$, where $X_l \in \mathbb{R}^{N \times D_l}$ stacks the activations and $\tilde X_l = X_l - \mathbf{1}\bar h_l^{\!\top}$.}
Let $P_d^{(l)}$ denote the orthogonal projector onto the top-$d$ principal directions of layer $l$.

\paragraph{Definition.}
For a threshold $\theta \in (0,1)$, the layer-$l$ intrinsic dimensionality is
\begin{equation}
  \dthetal \;:=\; \min\!\left\{ d \;:\; \mathbb{E}_{h \sim X_l}\!\left[\,\mathrm{cos}\bigl(h,\; {\hat h^{(d)}}\bigr)\,\right] \;\ge\; \theta \right\}{, \qquad \hat h^{(d)} \;=\; P_d^{(l)}\bigl(h - \bar h_l\bigr) + \bar h_l} .
  \label{eq:d-theta}
\end{equation}
{Because the principal subspace maximizes explained variance rather than expected cosine, $\dthetal$ counts the principal directions needed to reach $\theta$ and is therefore an upper bound on the dimension of the best $\theta$-accurate subspace.}
{In practice, $\dthetal$ is obtained by linearly interpolating the mean cosine between $d/D_l \in \{0.1, \ldots, 0.9\}$.}
We use $\theta = 0.99$ as the default threshold throughout the paper.
We treat $\theta$ as a hyperparameter that controls the resolution of the measurement and may be adjusted depending on the application.
The middle-heavy profile, calibration robustness, and pretraining-trajectory results reported in this paper are obtained at $\theta = 0.99${, and Appendix~\ref{app:theta-sweep} repeats the KD comparison at $\theta = 0.97$}.

\paragraph{Cosine similarity as the reconstruction metric.}
The Eckart--Young theorem~\citep{eckart1936approximation} makes top-$d$ PCA the Frobenius-optimal rank-$d$ approximation, which corresponds to maximizing explained variance.
However, the down-projection input $h_l$ contributes to the residual stream through $W_d^{(l)} h_l$, and the updated residual is renormalized at the next normalization step, which preserves direction while standardizing magnitude.
Information is therefore propagated primarily through direction, {which motivates a direction-based criterion over the variance-based one underlying PCA.
We accordingly define $\dthetal$ via a cosine threshold on the reconstruction $\hat h^{(d)}$ rather than by cumulative explained variance~\citep{jolliffe2002pca}.
The same spectrum can {also} be read out with an explained-variance threshold; \S\ref{subsec:five-way} and Appendix~\ref{app:ev-readout} compare the two.}

\paragraph{Layer-wise dispersion.}
To summarize a model's per-layer profile $(\dthetal/D_l)_{l}$ with a single scalar, we fix the cosine threshold to $\theta = 0.99$ and define
\begin{equation}
  \sigmas \;:=\; \mathrm{std}_l \!\left( d_{0.99,\,l} \,/\, D_l \right),
\end{equation}
the standard deviation of layer-relative $\dtheta$ across layers (evaluated at the default threshold).
We normalize by $D_l$ so that $\sigmas$ is comparable across architectures with different FFN intermediate dimensions.
Across the models we evaluate, $\sigmas$ ranges from {$\sim 0.06$} (nearly flat profiles) to $\sim 0.21$ (highly dispersed profiles), reflecting substantial variation in how non-uniformly capacity is allocated across depth.
$\sigmas$ serves as a compact descriptor of a model's structural heterogeneity and reappears in our analysis of post-training stability ({\S\ref{subsec:post-training-stability}}) and {in the KD validation, where it tracks the size of the gain (\S\ref{subsec:kd-predictors})}.

\section{What \texorpdfstring{$\dtheta$}{d-theta} Measures: Empirical Characterization}
\label{sec:what-d-measures}

We scanned 58 LLMs spanning 14 families and 70M--72B parameters with the protocol of \S\ref{sec:method}.
Four findings emerge.

\subsection{\texorpdfstring{The profile is robust to calibration data{, language, and measurement site}}{The profile is robust to calibration data, language, and measurement site}}
\label{subsec:cross-language}
\begin{figure}[t]
  \centering
  \includegraphics[width=0.95\linewidth]{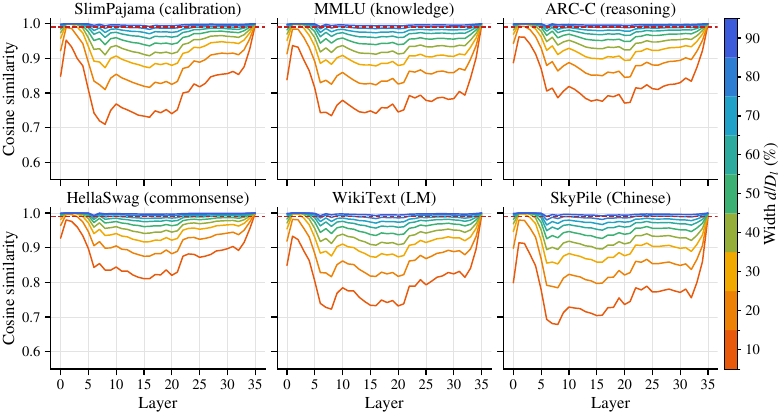}
  \caption{\textbf{Calibration robustness on Qwen3-8B.}
  Per-layer cosine similarity between $h_l$ and its rank-$d$ PCA reconstruction at width fractions $d/D_l \in \{10\%, 20\%, \ldots, 90\%\}$, measured on six 100K-token calibration corpora: five English (SlimPajama, MMLU, ARC-C, HellaSwag, WikiText) and Chinese SkyPile.
  Red dashed line marks the $\theta = 0.99$ threshold.
  All six panels show the same dip at middle layers when $d/D_l$ is small, indicating that more principal directions are required to faithfully reconstruct middle-layer activations.
  Pairwise per-layer Pearson $r$ on the extracted $\dthetal / D_l$ profiles: five English corpora min $0.953$, mean $0.975$; English $\leftrightarrow$ Chinese min $0.941$, mean $0.973$; all 15 pairs $\geq 0.941$, with the minimum occurring between Chinese SkyPile and HellaSwag.}
  \label{fig:protocol-overlay}
\end{figure}

We test whether $\dtheta$ depends on the choice of calibration corpus along two axes: distributional variation within a single language, and variation across languages.
On Qwen3-8B, we measure $\dtheta$ using six 100K-token corpora: five English (SlimPajama, MMLU, ARC-C, HellaSwag~\citep{zellers2019hellaswag}, WikiText~\citep{merity2017wikitext}) and one Chinese (SkyPile~\citep{wei2023skywork}).
Across the five English corpora, all pairwise per-layer Pearson correlations lie in $[0.953, 0.990]$ (mean $0.975$), indicating that $\dtheta$ is largely insensitive to the textual domain of the calibration data.
A stronger test is cross-language: per-layer correlations between Chinese SkyPile and each English corpus lie in $[0.941, 0.993]$ (mean $0.973$), comparable to the all-English pairwise correlations.
The lowest pairwise correlation across all 15 pairs is $0.941$ (Chinese SkyPile vs.\ HellaSwag), {reflecting} the outlier commonsense distribution of HellaSwag rather than the language gap.
The mean of $\dthetal / D_l$ shifts by $+0.064$ under Chinese calibration---a near-uniform offset that preserves the profile shape (Figure~\ref{fig:protocol-overlay}).
$\dtheta$ thus reflects a structural property of the weights rather than an artifact of the calibration distribution.

{The profile is also stable across measurement sites.
In addition to the down-projection input $h_l$, we compute the spectrum at four other points of Equation~\eqref{eq:gated-ffn} on every layer of six models from four families: the gate pre-activation $W_g^{(l)} x_l$, the gate post-activation $\phi(W_g^{(l)} x_l)$, the up-projection output $W_u^{(l)} x_l$, and the FFN output $W_d^{(l)} h_l$.
The FFN output, which is the quantity added to the residual stream, gives a middle-heavy profile in all six models that closely follows the $h_l$ profile (Pearson $r = 0.70$--$0.95$ in five models; the exception is OLMo-2-7B~\citep{olmo2_2024}, the lowest-$\sigmas$ model in the set, whose $h_l$ profile is not middle-heavy).
Only the gate post-activation departs from it (not middle-heavy in five of six models, anticorrelated in three), as expected for a quantity that enters the FFN only through its elementwise product with $W_u^{(l)} x_l$ and never reaches the residual stream.\par}

\subsection{Per-layer profile is recurringly middle-heavy}
\label{subsec:middle-heavy}

\begin{figure}[t]
  \centering
  \includegraphics[width=0.95\linewidth]{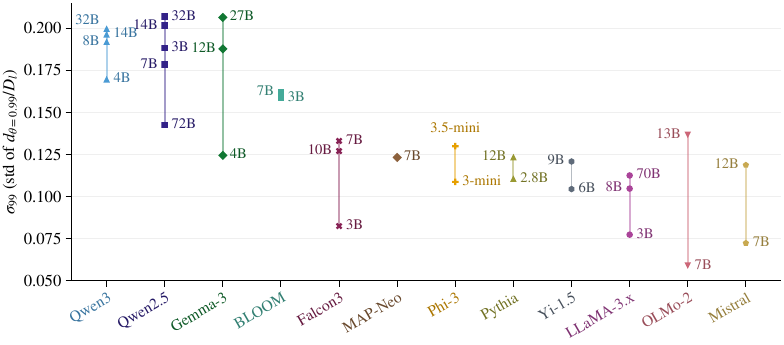}
  \caption{\textbf{{$\sigmas$ by model family.}}
  {Each column is one family, ordered by median $\sigmas$; markers are individual models of 2B parameters or more, labeled by size.}
  {Among the plotted models, $\sigmas$ shows a substantial association with model family.
  Within most families, $\sigmas$ also tends to rise with scale, although Qwen2.5 drops at 72B, the largest scanned model.}}
  \label{fig:sigma-vs-scale}
\end{figure}

Across the 58-model scan, the middle third of layers has higher mean $\dthetal / D_l$ than the edge thirds in {$76\%$ of the models (a mean-based criterion, not a strict bell shape)}.
{The exceptions} are concentrated in two regimes: small models ($\leq 1.7$B parameters; e.g., Qwen3-0.6B, LLaMA-3.2-1B, SmolLM2-135M/1.7B~\citep{allal2025smollm2}, BLOOM-560m/1b1~\citep{workshop2023bloom}, MAP-Neo-SL-250M~\citep{zhang2024mapneo}), and families with low $\sigmas$ ({OLMo-2~\citep{olmo2_2024}} 1B/7B/13B, LLaMA-3.x~\citep{grattafiori2024llama3} 1B/3B/8B, Mistral-7B/Nemo~\citep{jiang2023mistral,mistral2024nemo}); Figure~\ref{fig:sigma-vs-scale} shows the per-family $\sigmas$ distribution that motivates the latter regime.
{The shape is learned rather than architectural: a randomly initialized Qwen2.5-3B measured with the same protocol needs about $91\%$ of the width {on average} and is not middle-heavy, whereas the trained model shows the middle-heavy profile.}

\subsection{\texorpdfstring{$\sigmas$}{sigma99} trajectories show family-dependent dynamics during pretraining}
\label{subsec:multi-family-trajectory}

\begin{figure}[t]
  \centering
  \includegraphics[width=0.95\linewidth]{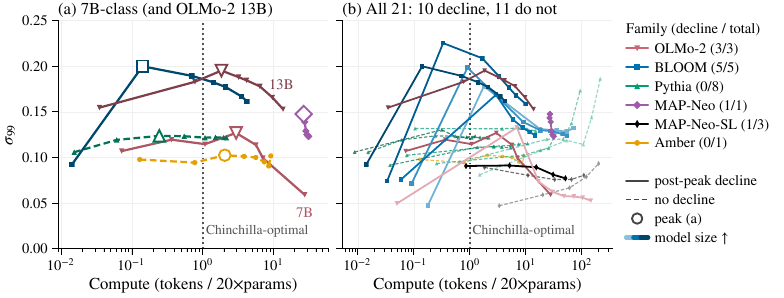}
  \caption{\textbf{$\sigmas$ trajectories across {six} families.}
  The horizontal axis is the Chinchilla-optimal compute multiplier: training tokens divided by $20\times$ the model's parameter count, where $20$ tokens-per-parameter is the compute-optimal ratio identified by~\citet{hoffmann2022chinchilla}.
  Values $<1$ indicate under-trained models relative to compute-optimal; values $>1$ indicate over-trained.
  Vertical dotted line marks the compute-optimal point.
  \textbf{(a)} 7B-class (and OLMo-2-13B) trajectories.
  {OLMo-2 and BLOOM show growth followed by sustained decline, and MAP-Neo-7B declines during its decay phase; Pythia-6.9b (terminating at $2.2\times$ Chinchilla) and Amber show no decline.}
  \textbf{(b)} All 21 trajectories.}
  \label{fig:multi-family-trajectory}
\end{figure}

Using publicly released checkpoints from pretraining runs across six families and two architectures, we track $\sigmas$ as a function of training compute (Figure~\ref{fig:multi-family-trajectory}).
The runs cover BLOOM~\citep{workshop2023bloom} (560M--7.1B, GELU), OLMo-2~\citep{olmo2_2024} (1B/7B/13B, SwiGLU), Pythia~\citep{biderman2023pythia} (70M--12B, GELU), Amber~\citep{liu2023llm360} (7B, SwiGLU), MAP-Neo~\citep{zhang2024mapneo} (7B, SwiGLU; {decay phase after 3.7T tokens}), and MAP-Neo-SL (250M/460M/980M, SwiGLU), for 21 trajectories in total, each with 5--8 checkpoints.
Compute is normalized as the Chinchilla-optimal~\citep{hoffmann2022chinchilla} multiplier $T / 20N$ ($T$ training tokens, $N$ parameters) to enable cross-scale comparison.

\textbf{{$\sigmas$ rises from the first checkpoint.}}
{Most trajectories exhibit a growth phase from the first checkpoint up to peak $\sigmas$ (20 of 21; the exception is MAP-Neo-SL-460M).
Peak values lie in $\sigmas \in [0.09, 0.23]$ depending on family and scale.}
The pattern holds across both gated SwiGLU and non-gated GELU FFNs.
{This indicates progressive layer differentiation: initially similar layers diverge in $\dtheta$ as training proceeds.}

\textbf{{Families share a profile shape.}}
{Within each family except BLOOM, the per-layer profiles have a similar shape (Figure~\ref{fig:per-layer-profiles} in Appendix~\ref{app:full-scan}).
$\sigmas$ also shows a substantial association with model family (family accounts for about half of its variance in the 58-model scan).}
At the 7B class, peak $\sigmas$ ranges from {$0.103$ (Amber) to $0.200$ (BLOOM-7b1), with MAP-Neo-7B, OLMo-2-7B, and Pythia-6.9b in between.}

\textbf{Post-peak behavior differs across families.}
Past the peak, $\sigmas$ declines {by more than $10\%$} in 10 trajectories, spanning OLMo-2, BLOOM, MAP-Neo, and MAP-Neo-SL families (Figure~\ref{fig:multi-family-trajectory}b).
Decline magnitudes range from {$-15\%$ (MAP-Neo-SL-980M) to $-60\%$, with the largest declines in OLMo-2-1B ($-60\%$ at $146\times$ Chinchilla) and OLMo-2-7B ($-53\%$ at $27.8\times$ Chinchilla).}
The remaining 11 trajectories (all 8 Pythia checkpoint sequences, Amber-7B, {and MAP-Neo-SL-250M/460M) end within $10\%$ of their peak.}
The decline-vs-no-decline distinction does not track total compute: small Pythia models trained at extremely high Chinchilla multipliers ($93$--$214\times$) show no decline, while several decline trajectories reach their peaks before Chinchilla-optimal and decline thereafter.
The trajectory shape thus depends on factors beyond total compute{, possibly including learning-rate (LR) schedule, data composition and architecture (\S\ref{sec:discussion})}.

\subsection{\texorpdfstring{$\sigmas$}{sigma99} is preserved through post-training}
\label{subsec:post-training-stability}

While pretraining substantially reshapes $\sigmas$ (\S\ref{subsec:multi-family-trajectory}), post-training does not.
We measured general {post-training (Qwen2.5-7B-Instruct~\citep{qwen2024qwen25}, Qwen3-8B~\citep{yang2025qwen3})}, domain-specific {post-training (Qwen2.5-Coder-7B-Instruct~\citep{hui2024qwen25coder}, Qwen2.5-Math-7B-Instruct~\citep{yang2024qwen25math})}, and mid-training annealing (OLMo-2~\citep{olmo2_2024} Stage 2).
In all cases $\sigmas$ changes by at most {$10\%$} {relative to its base value} ({Table~\ref{tab:post-training-sigma} in Appendix~\ref{app:full-scan}}), in contrast to the {$-53\%$ post-peak decline of OLMo-2-7B during Stage~1 pretraining (\S\ref{subsec:multi-family-trajectory})}.
{Domain-specific continued pretraining also leaves $\sigmas$ largely unchanged: Qwen2.5-Coder-7B and Qwen2.5-Math-7B, continued from Qwen2.5-7B on code and math corpora, differ from their base by $-0.1\%$ and $+4.8\%$.}
{This stability suggests that how unevenly intrinsic dimensionality is distributed across layers is largely determined during pretraining.}
Distillation checkpoints fall outside this regime.
DeepSeek-R1-Distill-Llama-8B~\citep{deepseek2025r1} has $\sigmas = {0.075}$, {$-26\%$} relative to its base {LLaMA-3.1-8B}~\citep{grattafiori2024llama3} ({$0.101$}); such procedures involve substantial weight rewriting and are out of scope for this paper.

\section{Validation: Controlled Width Pruning under KD}
\label{sec:kd-validation}

We treat the 3-way KD experiment as a validation of the measurement protocol.
The primary prediction is that \pca{}, with per-layer widths set to $d_{0.99}$, outperforms a budget-matched flat allocation \uni{}; an ablation at the tighter $\theta = 0.97$ budget is reported in Appendix~\ref{app:theta-sweep}.
We additionally include an inverted allocation \inv{} (the per-layer mirror of \pca{} around uniform) as a falsifiability check: a finding of \inv{} $\ge$ \uni{} would be consistent with the weaker hypothesis that any non-uniform allocation helps, while the observed \inv{} $<$ \uni{} {on all four teachers} identifies the shape of the allocation, not its non-uniformity, as the relevant signal.

\subsection{Setup}
\label{subsec:kd-setup}

\paragraph{Models.}
{We use four pretrained-only (base) teachers at moderate-to-large scale: Qwen2.5-3B, Qwen3-8B, Gemma-3-4B, and LLaMA-3-8B.}
For each teacher we distill three students under the same KD recipe.

\paragraph{Allocations.}
Let $L$ be the number of transformer layers and $D_l$ the teacher's FFN intermediate width at layer $l$.
For each teacher, we fix the student's total FFN parameter budget $B$ to the sum of the teacher's measured per-layer {$\dtheta$ values}: $B = \sum_l \dthetal$ at $\theta = 0.99$.
The budget is therefore not a tuning hyperparameter but a direct consequence of the threshold choice.

\begin{figure}[t]
  \centering
  \includegraphics[width=0.95\linewidth]{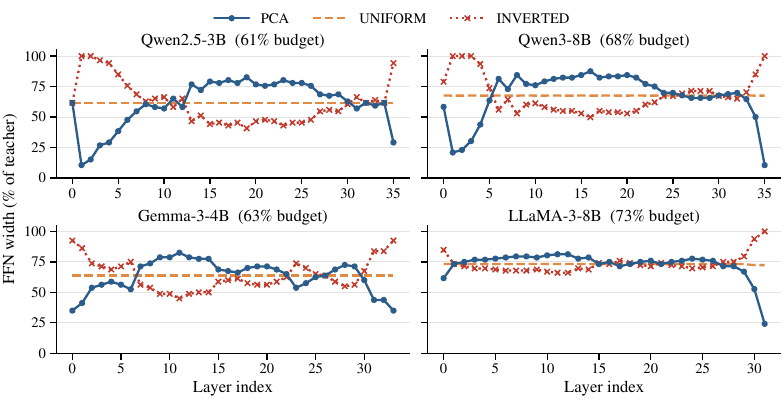}
  \caption{\textbf{Three width allocations under matched FFN budgets.}
  Per-layer FFN width as a percentage of the teacher's $D_l${; each title gives the total budget $\sum_l w_l / (L D)$}.
  \pca{} (blue, solid) follows $\dthetal$ at $\theta=0.99$; \uni{} (orange, dashed) is $B/L$; \inv{} (red, dotted) is the mirror of \pca{} around \uni{}, clipped at $D_l$.
  {Within each panel the three curves are budget-matched (\S\ref{subsec:kd-setup}).
  Visual dispersion across the curves scales with $\sigmas$: large in Qwen2.5-3B and Qwen3-8B ($\sigmas \approx 0.19$), smaller in Gemma-3-4B ($\sigmas = 0.124$) and LLaMA-3-8B ($\sigmas = 0.105$).}}
  \label{fig:3way-widths}
\end{figure}

Within budget $B$ we construct three layer-width vectors:
\begin{itemize}
  \item \pca: $w_l = \dthetal$ at $\theta = 0.99$, rounded to multiples of $128$.
  \item \uni: {$w_l \approx B/L$.}
  \item \inv: $w_l = 2\,\mathrm{uni}_l - \mathrm{pca}_l$ (the per-layer mirror of \pca{} around \uni{}), clipped to $[128, D_l]$ and rescaled to $B$.
\end{itemize}
{All three allocations match $B$ up to rounding, and any residual (at most $+1.3\%$, for \uni{} or \inv{}) favors the baseline;} Figure~\ref{fig:3way-widths} visualizes the per-layer width vectors for each teacher.

\paragraph{Initialization, distillation, and evaluation.}
We initialize each student via Minitron-style importance pruning~\citep{muralidharan2024compact}, scoring per-neuron importance by absolute activation magnitude over $128$ SlimPajama calibration sequences of length $2{,}048$, and retaining within each layer the top-scoring neurons up to that layer's allocated width.
We then distill against a pure Kullback--Leibler (KL) objective $\mathcal{L} = \mathrm{KL}(p_\text{teacher} \,\|\, p_\text{student})$ on SlimPajama at sequence length $2{,}048$, effective batch $256$ sequences, peak LR $1\!\cdot\!10^{-4}$ with a short warmup {and cosine decay (constant for Qwen3-8B; Appendix~\ref{app:per-task-per-step})}, 3K steps ($\approx 1.57$B tokens).
For evaluation we use five lm-eval-harness~\citep{evalharness} tasks (MMLU~\citep{hendrycks2021mmlu}, ARC-Challenge~\citep{clark2018arc}, HellaSwag~\citep{zellers2019hellaswag}, WinoGrande~\citep{sakaguchi2021winogrande}, PIQA~\citep{bisk2020piqa}) at harness-default few-shot settings, log-likelihood scoring without chat template, reporting the larger of \texttt{acc} and \texttt{acc\_norm} per task and the unweighted 5-task average as the headline number.

\subsection{\texorpdfstring{The 3-way ordering: \pca{} $>$ \uni{} $>$ \inv{}}{The 3-way ordering: PCA > Uniform > Inverted}}
\label{subsec:kd-ordering}

\paragraph{Monotonic ordering across {teachers}.}
{On all four teachers the three-way ordering \pca{} $>$ \uni{} $>$ \inv{} holds at step 3K on the 5-task average, and on Qwen2.5-3B every \pca{} run beats every \uni{} run, which beats every \inv{} run; it also holds at the $8$K-step horizon on the three teachers trained that long (Appendices~\ref{app:per-task-per-step} and~\ref{app:multi-seed}).
At the tighter $\theta = 0.97$ budget ($42.9\%$ of the teacher's FFN parameters), the advantage over \uni{} is $+2.86$\pp{}, no smaller than at $\theta = 0.99$ (Appendix~\ref{app:theta-sweep}).
Table~\ref{tab:3way-headline} reports headline numbers at step 3K and Table~\ref{tab:per-task-qwen25-3b} the per-task breakdown for Qwen2.5-3B, where \pca{} achieves the highest score on all five tasks, with the largest margin on MMLU ($+4.71$\pp{} vs.\ \uni) and the smallest on PIQA ($+0.80$\pp).}
{The ordering also holds without a teacher (\S\ref{subsec:lm-only}).}

\begin{table}[t]
\centering
\small
\setlength{\tabcolsep}{5pt}
\caption{\textbf{3-way KD headline results at step 3K.}
5-task averages (MMLU, ARC-C, HellaSwag, WinoGrande, PIQA).
Teacher avg is the unpruned model's same-protocol score.
{Rows are ordered by $\sigmas$.}
{Qwen2.5-3B and LLaMA-3-8B: means over three training seeds.
}}
\label{tab:3way-headline}
\begin{tabular}{lccccc}
\toprule
Model & $\sigmas$ & \pca{} avg & $\Delta$ vs.\ \uni & $\Delta$ vs.\ \inv & Teacher avg \\
\midrule
Qwen3-8B      & 0.192 & 67.48 & $+2.28$ & $+4.63$ & 72.23 \\
Qwen2.5-3B    & 0.188 & 60.10 & $+2.50$ & $+5.52$ & 66.49 \\
Gemma-3-4B    & 0.124 & 54.24 & $+0.88$ & $+2.05$ & 67.56 \\
LLaMA-3-8B    & 0.105 & 61.83 & $+0.52$ & $+0.99$ & 67.42 \\
\bottomrule
\end{tabular}

\end{table}

\begin{table}[t]
\centering
\small
\caption{{\textbf{Per-task results for Qwen2.5-3B at step 3K}, means over three training seeds per allocation.
The $\Delta(\pca-\inv)$ column is led by MMLU, where the inverted allocation reaches $37.93$ against $52.25$ for \pca{}.}}
\label{tab:per-task-qwen25-3b}
\begin{tabular}{lrrrrr}
\toprule
Task & \pca & \uni & \inv & $\Delta(\pca - \uni)$ & $\Delta(\pca - \inv)$ \\
\midrule
MMLU       & \textbf{52.25} & 47.54 & 37.93 & $+4.71$ & $+14.32$ \\
ARC-C      & \textbf{42.01} & 41.18 & 38.99 & $+0.82$ & $+3.01$ \\
HellaSwag  & \textbf{65.77} & 63.79 & 61.16 & $+1.98$ & $+4.61$ \\
WinoGrande & \textbf{64.85} & 60.64 & 60.04 & $+4.21$ & $+4.81$ \\
PIQA       & \textbf{75.61} & 74.81 & 74.77 & $+0.80$ & $+0.83$ \\
\midrule
\textbf{Avg} & \textbf{60.10} & 57.59 & 54.58 & $+2.50$ & $+5.52$ \\
\bottomrule
\end{tabular}

\end{table}

\paragraph{Init-vs-trained reversal rules out a static init-advantage account.}
{At step $0$, before any KD, the ordering on Qwen3-8B is \uni{} ($51.47$) $>$ \pca{} ($49.92$) $>$ \inv{} ($47.84$).}
After 1K steps the ordering reverses to \pca{} ($66.10$) $>$ \uni{} ($63.05$) $>$ \inv{} ($60.35$) and persists at steps 2K and 3K.
{We observe the same pattern on LLaMA-3-8B, where \uni{} leads at init and \pca{} leads the three-seed mean at step 3K and the single longer run at steps 5K and 8K (Table~\ref{tab:per-step-3way} in Appendix~\ref{app:per-task-per-step}).}
{On Qwen3-8B and LLaMA-3-8B, \pca{}'s advantage therefore does not come from a better post-pruning initialization; it arises during KD training.}

\begingroup
\subsection{Comparison with other allocation rules}
\label{subsec:five-way}

We compare \pca{} with two other budget-matched rules on Qwen2.5-3B and Gemma-3-4B (Table~\ref{tab:five-way}).
\ev{} reads the same spectrum with the cumulative explained-variance rule~\citep{jolliffe2002pca}, the criterion behind layer saturation~\citep{shenk2019spectral,richter2021saturation}; \slim{} applies the layer-ratio rule of \citet{guo2025slimllm} ($\alpha = 8$), a redundancy signal based on the cosine similarity between each layer's input and output.
{\ev{} is comparable to \pca{} on Qwen2.5-3B but falls below it on Gemma-3-4B, and \slim{} trails \pca{} on both teachers, most on MMLU.
The two readouts differ most on layers dominated by a single direction (Figure~\ref{fig:ev-widths}), where the explained-variance rule assigns near-minimal width; this has no measurable cost on Qwen2.5-3B (one early layer) but hurts on Gemma-3-4B (two mid-depth layers).
The cosine readout keeps width on such layers and appears more robust across the two teachers.}

{The rules above all read a profile from the model being pruned; a further question is whether any sufficiently middle-heavy profile would do.
Transferring the more strongly middle-heavy profile measured on Qwen3-8B onto LLaMA-3-8B at the same budget gives $60.53$ at step 3K, below \uni{} ($60.85$) and level with \inv{} ($60.48$), whereas LLaMA-3-8B's own \pca{} profile reaches $61.72$ (single runs; Appendix~\ref{app:controls}).
This supports using the model's own measured profile rather than a generic middle-heavy allocation.}

\begin{table}[t]
\centering
\small
\setlength{\tabcolsep}{5pt}
\caption{\textbf{\pca{} against two other allocation rules at step 3K.}
Qwen2.5-3B: mean $\pm$ SD over three training seeds per allocation.
\uni{} scores $57.59$ on Qwen2.5-3B and $53.36$ on Gemma-3-4B (Table~\ref{tab:per-step-3way}).}
\label{tab:five-way}
\begin{tabular}{llccc}
\toprule
Teacher & Allocation & 5-task avg & MMLU & $\Delta$ (\pca{} $-$ alloc.) \\
\midrule
\multirow{3}{*}{Qwen2.5-3B} & \pca  & $60.10 \pm 0.22$          & 52.25          & --- \\
                            & \ev   & $\mathbf{60.18} \pm 0.10$ & \textbf{52.56} & $-0.08$ \\
                            & \slim & $56.69 \pm 0.05$          & 44.62          & $+3.41$ \\
\midrule
\multirow{3}{*}{Gemma-3-4B} & \pca  & $\mathbf{54.24}$          & \textbf{42.59} & --- \\
                            & \ev   & $52.60$                   & 38.76          & $+1.64$ \\
                            & \slim & $52.88$                   & 38.75          & $+1.36$ \\
\bottomrule
\end{tabular}
\end{table}
\endgroup

\subsection{\texorpdfstring{{$\sigmas$ tracks the size of the gain}}{sigma99 tracks the size of the gain}}
\label{subsec:kd-predictors}

{The 3-way ordering holds on each of the four teachers, but its magnitude varies, and the variation tracks $\sigmas$ (Table~\ref{tab:3way-headline}).
On the two high-$\sigmas$ teachers (Qwen2.5-3B, $0.188$; Qwen3-8B, $0.192$), $\Delta(\pca{}-\uni{})$ is $+2.50$ and $+2.28$\pp{} and $\Delta(\pca{}-\inv{})$ is $+5.52$ and $+4.63$\pp{}; on the lowest-$\sigmas$ teacher, LLaMA-3-8B ($0.105$), the gaps shrink to $+0.52 \pm 0.26$ and $+0.99 \pm 0.46$\pp{}, still positive in every seed.
The gaps differ far more than $\sigmas$ does: when the profile is more dispersed, following it helps more and inverting it costs more.}

{Scale alone does not account for the gap.
LLaMA-3-8B matches Qwen3-8B in size but shows a much smaller gap.
The 1B-class teachers (LLaMA-3.2-1B, Qwen2.5-1.5B, Gemma-3-1B) also give $\Delta(\pca{}-\uni{}) < 0.6$\pp{} (Appendix~\ref{app:1b-class}), but they likewise have low $\sigmas$, so size and dispersion cannot be separated at that scale.
Every small-gap teacher, including Gemma-3-4B, has a lower $\sigmas$ than either large-gap teacher.
Four teachers show a trend rather than a functional form, and $\sigmas$ is confounded with family and recipe (\S\ref{sec:discussion}).}

\begingroup
\subsection{The ordering holds without distillation}
\label{subsec:lm-only}

To test whether the ordering depends on the teacher, we train the Minitron-pruned Qwen2.5-3B students with the language-modeling cross-entropy in place of the KD loss, keeping the data, the $3$K steps, the three allocations and the Qwen2.5-3B KD learning-rate recipe (\S\ref{subsec:kd-setup}).
The ordering is the same as under KD: \pca{} $>$ \uni{} $>$ \inv{} holds at every checkpoint from 1K to 3K, with gaps of $+2.31$ and $+5.63$\pp{} at step 3K ($56.69$, $54.37$ and $51.06$; gaps from unrounded values).
The advantage of the measured allocation is therefore not specific to teacher supervision.
\endgroup

\subsection{The advantage persists through SFT}
\label{subsec:kd-sft}

To test whether the KD-time advantage survives a downstream instruction-tuning stage, we applied identical SFT to the three Qwen3-8B step-3K students (Table~\ref{tab:sft-downstream}).
SFT setup: tulu-3-100k (a $100$K-example stratified subset of \texttt{tulu-3}~\citep{lambert2024tulu} preserving the original category proportions), 2 epochs, batch $128$, learning rate $2 \times 10^{-5}$.
We evaluate on \texttt{arc\_challenge\_chat}~\citep{clark2018arc}, \texttt{mmlu\_redux\_generative}~\citep{gema2024mmluredux}, and GSM8K~\citep{cobbe2021gsm8k} from lm-eval-harness{, zero-shot with the chat template}.
{After SFT, \pca{} leads \uni{} by $+11.0$, $+4.9$ and $+3.0$\pp{} on ARC-C, MMLU and GSM8K and leads \inv{} by $+10.3$ to $+11.8$\pp;} the \pca{} $>$ \uni{} $>$ \inv{} ordering also holds at each individual epoch.

\begin{table}[t]
\centering
\small
\caption{\textbf{Qwen3-8B step-3K students after SFT (\texttt{tulu-3-100k}, 2 epochs, best-of-epoch).}
\pca{} $>$ \uni{} $>$ \inv{} on every task.}
\label{tab:sft-downstream}
\begin{tabular}{lrrr}
\toprule
Config & ARC-C & MMLU & GSM8K \\
\midrule
\pca       & \textbf{85.58} & \textbf{67.75} & \textbf{73.46} \\
\uni       & 74.57 & 62.83 & 70.43 \\
\inv       & 73.81 & 57.41 & 62.55 \\
\midrule
$\Delta(\pca{}-\uni{})$ & $+11.01$ & $+4.92$ & $+3.03$ \\
$\Delta(\pca{}-\inv{})$ & $+11.77$ & $+10.34$ & $+10.91$ \\
\bottomrule
\end{tabular}
\end{table}

The width-allocation choice influences a property that survives a 2-epoch SFT round on a fresh dataset, so it is not a transient KD-time artifact.
This is a deployment-relevant signal: in production pipelines that combine KD-pruning with downstream SFT, the choice of per-layer width carries through to user-facing benchmark performance.

\section{Discussion and Limitations}
\label{sec:discussion}

\paragraph{What drives the family-level $\sigmas$ signatures?}
{$\sigmas$ shows a substantial association with model family} and spans more than $2\times$ across families (Figure~\ref{fig:sigma-vs-scale}: {$0.18$--$0.21$ for Qwen2.5 3B--32B and Qwen3 8B--32B} vs.\ LLaMA $0.08$--$0.11$).
{Identifying the responsible recipe ingredient would require iso-architecture iso-data ablations that public checkpoints cannot provide, and the same limitation restricts the $\sigmas$-vs-gap pattern to the four teachers tested ($\sigmas \in [0.105, 0.192]$); iso-family iso-scale variants across a wider $\sigmas$ range would help disentangle recipe from $\sigmas$.}

\begingroup
\paragraph{Limitations.}
Two limitations concern scale and deployment.
Our runs cover $1.57$--$4.2$B training tokens; whether the advantage persists at Minitron-scale budgets ($70$--$100$B) is untested.
A non-uniform student also carries a serving cost: in our vLLM~\citep{kwon2023vllm} integration the trained \pca{} student is $5.3\%$ slower than \uni{} at a $0.4\%$ parameter difference (Appendix~\ref{app:serving}).
\endgroup

\paragraph{Toward {$\dtheta$-guided} width re-allocation beyond KD.}
{Two natural extensions are re-shaping the same pretrained model under a preserved budget with continued pretraining, and selective capacity augmentation at high-$\dtheta$ layers of an already-trained model, motivated by the observation that $\sigmas$ changes little after pretraining (\S\ref{subsec:post-training-stability}).
Whether the advantage persists under either scheme is the top future ablation.
A layer-output perturbation or pruning-sensitivity study would further test whether high-$\dtheta$ layers are also more capacity-sensitive.}

\section{Related Work}
\label{sec:related}

\paragraph{Functional heterogeneity of FFN layers.}
FFN layers act as key-value memories~\citep{geva2021transformer} whose updates promote vocabulary concepts~\citep{geva2022ffn}; causal tracing localizes factual associations to middle multi-layer perceptrons (MLPs)~\citep{meng2022rome}, and \citet{dai2022knowledge} identify knowledge neurons in BERT~\citep{devlin2019bert}.
These studies suggest that FFN layers at different depths perform qualitatively different functions; our $\dtheta$ provides a complementary, quantitative signal of how many activation directions each layer uses.

\paragraph{Width-aware compression and pruning.}
\textsc{Wanda}~\citep{sun2024wanda} and SparseGPT~\citep{frantar2023sparsegpt} induce unstructured sparsity, and {LLM-Pruner~\citep{ma2024llm}, Sheared LLaMA~\citep{xia2023sheared}, ShortGPT~\citep{gong2024shortgpt}, SliceGPT~\citep{ashkboos2024slicegpt} and Olica~\citep{he2025olica} remove neuron groups, heads, layers or dimensions; none of them allocates per-layer FFN width from a measured intrinsic-dimensionality profile under a fixed budget.
SlimLLM~\citep{guo2025slimllm} does allocate a per-layer ratio, and we compare against its rule in \S\ref{subsec:five-way}.
W-PCA~\citep{wang2025wpca} counts PCA components of FFN states in untrained BERT-scale networks as a zero-shot neural architecture search (NAS) proxy; we measure trained LLMs and use the profile directly as the allocation.
Minitron~\citep{muralidharan2024compact} couples width pruning with distillation under a uniform ratio; we adopt its recipe and replace the uniform allocation with $\dtheta$.}
Closest in motivation, \citet{ikeda2025layerwise} report a binary middle-heavy FFN allocation trained from scratch on a single architecture family at $285$M--$1.2$B parameters; we differ in (i) producing a {non-binary} $\dtheta$ per layer and (ii) validating with a controlled KD experiment that isolates width allocation from training procedure.

\paragraph{Effective rank and intrinsic dimensionality.}
Effective rank~\citep{roy2007effective} and nonlinear intrinsic-dimension estimators~\citep{ansuini2019intrinsic} characterize layer representations; \citet{aghajanyan2021intrinsic} instead use a parameter-space intrinsic dimension to explain fine-tuning.
\citet{ansuini2019intrinsic} report a ``hunchback'' profile in convolutional neural networks (CNNs), in which intrinsic dimensionality first increases and then progressively decreases toward the output, and note that linear estimates such as PCA do not recover it; our PCA-based measure recovers an analogous middle-heavy profile in LLM FFNs.
{\citet{shenk2019spectral} and \citet{richter2021saturation} define layer ``saturation'' as the share of dimensions needed to reach a fixed cumulative explained variance and use it to diagnose over-provisioned layers during training.}
Our $\dtheta$ is in this family but specialized to FFN down-projection input activations, and uses a cosine-similarity threshold rather than explained variance, {motivated by the direction-additive way in which FFN outputs enter the residual stream; it separates from saturation on layers dominated by a single direction, where the cosine readout retains substantially more width (\S\ref{subsec:five-way}, Appendix~\ref{app:ev-readout})}.

\section{Conclusion}
\label{sec:conclusion}

We define $\dtheta$ via cosine-thresholded PCA on the FFN down-projection input and document a recurring middle-heavy, calibration-robust, post-training-stable profile across 58 LLMs spanning 14 families from 70M to 72B parameters.
The single-number dispersion $\sigmas$ derived from the profile is a model-internal property that survives the shift from English to Chinese calibration and {changes little} across post-training fine-tuning.
We validate the signal in a 3-way KD experiment in which budget-matched \pca-guided allocation beats \uni{} on {four} teachers and a budget-matched \inv{} control {does worst}, with the magnitude of the advantage tracking $\sigmas$ across teachers and persisting through downstream supervised fine-tuning.

\section*{{Acknowledgments}}
\leavevmode{We thank Geewook Kim for helpful discussions during the review process and for proofreading the manuscript.}

\bibliographystyle{plainnat}
\bibliography{references}

\clearpage
\appendix

\section{\texorpdfstring{$\theta$}{theta} Sweep}
\label{app:theta-sweep}

\paragraph{Higher-compression $\theta = 0.97$ ablation on Qwen2.5-3B.}
The body experiments use the cosine threshold $\theta = 0.99$, which on Qwen2.5-3B yields a $61.3\%$ FFN-parameter budget ($\sum_l \dthetal = 242{,}816$ relative to the teacher's $L \cdot D = 396{,}288$).
To test whether the headline ordering depends on the size of that budget, we re-run the 3-way KD recipe at $\theta = 0.97$, which produces a tighter PCA allocation summing to $169{,}984$ FFN-intermediate units ($42.9\%$ of the teacher budget, $1.43\times$ smaller than at $\theta = 0.99$).
The \uni{} and \inv{} controls are budget-matched to this $\theta = 0.97$ \pca{} sum so the comparison isolates the shape of the allocation while holding total FFN parameters fixed.
All other hyperparameters (teacher, optimizer, KD loss, batch size, $3{,}000$-step horizon, evaluation suite) are identical to the body $\theta = 0.99$ runs{, including the learning-rate recipe (\S\ref{subsec:kd-setup})}.

Table~\ref{tab:theta-097-qwen25-3b} reports the 5-task average at every evaluated KD step.
The \pca{} $>$ \uni{} $>$ \inv{} ordering holds at every post-init step{, as for the $\theta=0.99$ runs in Table~\ref{tab:per-step-3way}}.
{At step 3K the gap to \uni{} does not shrink with the smaller budget ($\Delta(\pca{}-\uni{}) = +2.86$\pp{} at $\theta=0.97$ against $+2.50$\pp{} at $\theta=0.99$, Table~\ref{tab:3way-headline}), while the gap to \inv{} narrows ($+4.46$\pp{} against $+5.52$\pp).
The $\theta=0.97$ sweep is a single training run per allocation; the matching single $\theta=0.99$ runs give $+2.44$ and $+5.47$\pp, so the comparison is the same against them.}

\begin{table}[h]
\centering
\small
\setlength{\tabcolsep}{4pt}
\caption{\textbf{Qwen2.5-3B 3-way KD at $\theta = 0.97$ (5-task average).}
Bold = best per step{; task scores follow the body's protocol (\S\ref{subsec:kd-setup})}.
The PCA allocation sums to $169{,}984$ FFN units ($42.9\%$ of the teacher's $L \cdot D = 396{,}288$), a $1.43\times$ smaller budget than the body's $\theta=0.99$ {runs} ($61.3\%$ budget).
\uni{} and \inv{} are budget-matched to the same $\theta = 0.97$ sum.
The \pca{} $>$ \uni{} $>$ \inv{} ordering holds at every post-init step, matching the $\theta=0.99$ post-init ordering on the same teacher in Table~\ref{tab:per-step-3way}.}
\label{tab:theta-097-qwen25-3b}
\begin{tabular}{lccc}
\toprule
Config & step 1K & step 2K & step 3K \\
\midrule
\pca       & $\mathbf{51.22}$ & $\mathbf{52.71}$ & $\mathbf{53.42}$ \\
\uni       & $48.73$          & $50.23$          & $50.56$ \\
\inv       & $46.81$          & $48.32$          & $48.96$ \\
\midrule
$\Delta(\pca{}-\uni{})$ & $+2.49$ & $+2.48$ & $+2.86$ \\
$\Delta(\pca{}-\inv{})$ & $+4.41$ & $+4.39$ & $+4.46$ \\
\bottomrule
\end{tabular}

\end{table}

\section{58-Model PCA Scan Summary}
\label{app:full-scan}

The full 58-model scan covers 14 model families ranging from 70M (Pythia-70m) to 72B (Qwen2.5-72B) parameters.
For each model we extract parameter count, hidden dimension, FFN intermediate width $D$, number of transformer layers $L$, {the dispersion descriptor $\sigmas$ (standard deviation of $\dthetal/D_l$ at $\theta=0.99$), the saturation ratio (mean $\dthetal/D_l$), the prunable fraction ($1$ minus the saturation ratio), and the mean $\dthetal/D_l$ of the first, middle and last thirds of layers}.
The summary statistics in the body (\S\ref{sec:what-d-measures}) are computed over this set, and the per-family $\sigmas$ distribution and per-layer profiles are visualized in Figure~\ref{fig:sigma-vs-scale} (\S\ref{subsec:middle-heavy}) and Figure~\ref{fig:per-layer-profiles} below {(profiles of all 58 models are in the supplementary material)}; pretraining trajectories are visualized in Figure~\ref{fig:multi-family-trajectory} (\S\ref{subsec:multi-family-trajectory}).
{Sixteen of the 58 models (BLOOM-560m/1b1/1b7/3b/7b1, Pythia-70m/160m/410m/1b/1.4b/2.8b/12b, MAP-Neo-7B {(end of its decay phase)} and MAP-Neo-SL-250M/460M/980M) are the final checkpoints of the pretraining trajectories in \S\ref{subsec:multi-family-trajectory}, measured with the same protocol as the other models.}

\begin{figure}[t]
  \centering
  \includegraphics[width=0.98\linewidth]{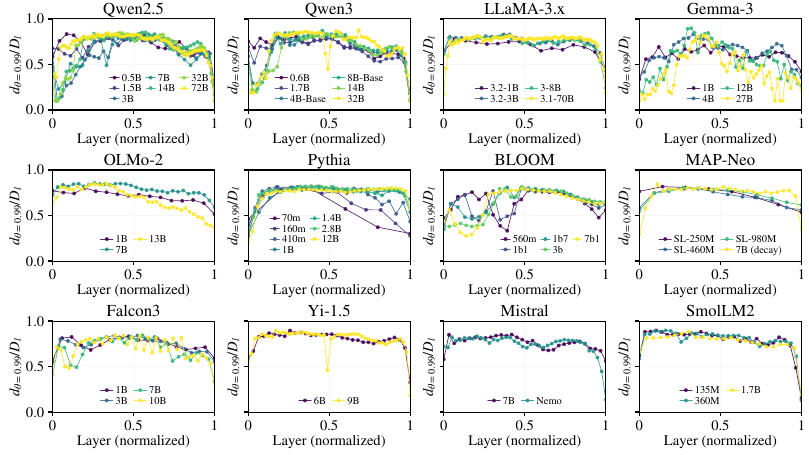}
  \caption{\textbf{Per-layer $\dthetal/D_l$ profiles for {51 of the 58 models} in the scan, grouped by family.}
  {Profiles of all 58 models are in the supplementary material.}
  Horizontal axis is normalized layer index; vertical axis is $\dthetal/D_l$ at $\theta = 0.99$.
  The middle-heavy profile recurs across architecture, training corpus, {and scale}.
  How strongly middle-heavy it is varies by family (Qwen2.5/Qwen3 most pronounced, LLaMA-3.x, Mistral, and Yi-1.5~\citep{young2024yi} flatter); checkpoints that are visibly not middle-heavy are concentrated in the $\le 1.7$B-parameter regime and the low-$\sigmas$ families flagged in the body.}
  \label{fig:per-layer-profiles}
\end{figure}

\begin{table}[h]
\centering
\small
\caption{\textbf{Post-training $\sigmas$ stability.}
{Across general and domain-specific post-training, domain-specific continued pretraining (CPT), and mid-training annealing,} $\sigmas$ changes by at most {$10\%$} {relative to its base model; $\Delta$ is computed from unrounded values}.
{$\Delta$ is relative to the row's base: Qwen2.5-7B for Qwen2.5-7B-Instruct and the two CPT rows, and Qwen2.5-Coder-7B / Qwen2.5-Math-7B for their -Instruct rows.}
This contrasts with the {$-53\%$ post-peak decline of OLMo-2-7B during Stage~1 pretraining} (Figure~\ref{fig:multi-family-trajectory}).}
\label{tab:post-training-sigma}
\begin{tabular}{llrr}
\toprule
Model & Stage & $\sigmas$ & $\Delta$ vs.\ base \\
\midrule
Qwen2.5-7B                & pretrain only              & 0.178 & --- \\
Qwen2.5-7B-Instruct       & + post-training            & 0.176 & $-1.5\%$ \\
Qwen2.5-Coder-7B          & + code CPT                 & 0.178 & $-0.1\%$ \\
Qwen2.5-Coder-7B-Instruct & + code CPT + post-training & 0.161 & $-9.4\%$ \\
Qwen2.5-Math-7B           & + math CPT                 & 0.187 & $+4.8\%$ \\
Qwen2.5-Math-7B-Instruct  & + math CPT + post-training & 0.185 & $-1.0\%$ \\
Qwen3-8B (base) & pretrain only       & 0.192 & --- \\
Qwen3-8B (post) & + post-training     & 0.183 & $-4.5\%$ \\
OLMo-2-7B Stage~1   & pretrain (final)    & 0.059 & --- \\
OLMo-2-7B main      & + 50B Stage~2 anneal & 0.059 & $-0.6\%$ \\
\bottomrule
\end{tabular}

\end{table}

\section{\texorpdfstring{Per-Step KD Results Across {Four} Teachers}{Per-Step KD Results Across Four Teachers}}
\label{app:per-task-per-step}

{All KD runs use AdamW~\citep{loshchilov2019adamw} ($\beta_1 = 0.9$, $\beta_2 = 0.95$, weight decay $0.1$) with gradient clipping at $1.0$ and peak LR $1\!\cdot\!10^{-4}$. Qwen2.5-3B and Gemma-3-4B use warmup $300$ and cosine decay to $1\!\cdot\!10^{-5}$ over an $8$K-step schedule, LLaMA-3-8B warmup $500$ and cosine decay to $1\!\cdot\!10^{-6}$ over $8$K steps, and Qwen3-8B warmup $300$ and a constant learning rate thereafter; results are reported at step 3K unless noted.}

{Table~\ref{tab:per-step-3way} reports the 5-task average at init and at KD steps 3K, 5K and 8K for each of the four teachers and three width allocations.
Two patterns are visible: (i) in the reported values the \pca{} $>$ \uni{} $>$ \inv{} ordering holds at every post-init step on every teacher, while at step $0$ (before any KD training) the cold-start best is \uni{} on Qwen3-8B and LLaMA-3-8B and \pca{} on Gemma-3-4B and Qwen2.5-3B;} {(ii) at every step the gaps are larger on the high-$\sigmas$ teachers than on LLaMA-3-8B; on Qwen3-8B $\Delta(\pca-\uni)$ narrows slightly from $+3.04$\pp{} at step 1K (\S\ref{subsec:kd-ordering}; from unrounded values) to $+2.28$\pp{} at step 3K, while on Qwen2.5-3B $\Delta(\pca-\uni)$ and $\Delta(\pca-\inv)$ are $+2.50$ and $+5.52$\pp{} at step 3K and, in the single longer run, $+2.28$ and $+5.18$\pp{} at step 5K and $+2.16$ and $+5.03$\pp{} at step 8K.}

\begin{table}[t]
\centering
\small
\setlength{\tabcolsep}{4pt}
\caption{{\textbf{5-task average at init and at KD steps 3K, 5K and 8K for all four teachers $\times$ three width allocations.}
Init values are the post-pruning, pre-KD checkpoints, shared across runs.
Step-3K entries for Qwen2.5-3B and LLaMA-3-8B are means over three training seeds.
At every post-init step shown the ordering is \pca{} $>$ \uni{} $>$ \inv{} on every teacher.}}
\label{tab:per-step-3way}
\begin{tabular}{lccccc}
\toprule
Teacher ($\sigmas$) & Config & init & step 3K & step 5K & step 8K \\
\midrule
\multirow{3}{*}{Qwen2.5-3B (0.188)}
 & \pca       & $\mathbf{44.06}$ & $\mathbf{60.10}$ & $\mathbf{60.34}$ & $\mathbf{60.22}$ \\
 & \uni       & $43.02$ & $57.59$ & $58.06$ & $58.06$ \\
 & \inv       & $40.97$ & $54.58$ & $55.16$ & $55.19$ \\
\midrule
\multirow{3}{*}{Qwen3-8B (0.192)}
 & \pca       & $49.92$ & $\mathbf{67.48}$ & --- & --- \\
 & \uni       & $\mathbf{51.47}$ & $65.20$ & --- & --- \\
 & \inv       & $47.84$ & $62.85$ & --- & --- \\
\midrule
\multirow{3}{*}{Gemma-3-4B (0.124)}
 & \pca  & $\mathbf{39.41}$ & $\mathbf{54.24}$ & $\mathbf{54.82}$ & $\mathbf{55.26}$ \\
 & \uni  & $39.21$ & $53.36$ & $53.63$ & $54.23$ \\
 & \inv  & $38.91$ & $52.19$ & $52.92$ & $53.17$ \\
\midrule
\multirow{3}{*}{LLaMA-3-8B (0.105)}
 & \pca       & $55.10$ & $\mathbf{61.83}$ & $\mathbf{63.49}$ & $\mathbf{64.22}$ \\
 & \uni       & $\mathbf{56.86}$ & $61.31$ & $62.71$ & $63.42$ \\
 & \inv       & $55.82$ & $60.84$ & $62.40$ & $62.54$ \\
\bottomrule
\end{tabular}

\end{table}

Tables~\ref{tab:per-task-qwen3-8b} and~\ref{tab:per-task-llama3-8b} report the per-task breakdown for Qwen3-8B and LLaMA-3-8B at step 3K.

\begin{table}[t]
\centering
\small
\caption{\textbf{Per-task results for Qwen3-8B at step 3K.}
\pca{} achieves the highest score on four of five tasks; \uni{} narrowly leads on PIQA.}
\label{tab:per-task-qwen3-8b}
\begin{tabular}{lrrrrr}
\toprule
Task & \pca & \uni & \inv & $\Delta(\pca - \uni)$ & $\Delta(\pca - \inv)$ \\
\midrule
MMLU       & \textbf{65.69} & 60.94 & 56.79 & $+4.75$ & $+8.90$ \\
ARC-C      & \textbf{52.73} & 49.06 & 46.59 & $+3.67$ & $+6.14$ \\
HellaSwag  & \textbf{72.33} & 71.23 & 69.40 & $+1.10$ & $+2.93$ \\
WinoGrande & \textbf{69.85} & 67.25 & 65.51 & $+2.60$ & $+4.34$ \\
PIQA       & 76.82          & \textbf{77.53} & 75.95 & $-0.71$ & $+0.87$ \\
\midrule
\textbf{Avg} & \textbf{67.48} & 65.20 & 62.85 & $+2.28$ & $+4.63$ \\
\bottomrule
\end{tabular}
\end{table}

\begin{table}[t]
\centering
\small
\caption{\textbf{Per-task results for LLaMA-3-8B at step 3K}{, means over three training seeds}.
\pca{} achieves the highest score on four of five tasks; \uni{} narrowly leads on PIQA.
{The per-task gaps are smaller on average than on the high-$\sigmas$ teachers, and $\Delta(\pca-\inv)$ is smaller on every task}, consistent with the {$\sigmas$ trend in \S\ref{subsec:kd-predictors}}.}
\label{tab:per-task-llama3-8b}
\begin{tabular}{lrrrrr}
\toprule
Task & \pca & \uni & \inv & $\Delta(\pca - \uni)$ & $\Delta(\pca - \inv)$ \\
\midrule
MMLU       & \textbf{44.11} & 43.15 & 41.62 & $+0.96$ & $+2.49$ \\
ARC-C      & \textbf{43.63} & 43.46 & 42.75 & $+0.17$ & $+0.88$ \\
HellaSwag  & \textbf{74.16} & 73.58 & 73.50 & $+0.58$ & $+0.66$ \\
WinoGrande & \textbf{69.01} & 67.90 & 68.22 & $+1.11$ & $+0.79$ \\
PIQA       & 78.22          & \textbf{78.45} & 78.11 & $-0.23$ & $+0.11$ \\
\midrule
\textbf{Avg} & \textbf{61.83} & 61.31 & 60.84 & $+0.52$ & $+0.99$ \\
\bottomrule
\end{tabular}
\end{table}

\section{1B-Class Scale Ablation}
\label{app:1b-class}

We extend the 3-way KD validation of \S\ref{sec:kd-validation} to three 1B-class teachers: Qwen2.5-1.5B~\citep{qwen2024qwen25}, LLaMA-3.2-1B~\citep{grattafiori2024llama3}, and Gemma-3-1B~\citep{team2025gemma3}.
All three are low-$\sigmas$ teachers ($\sigmas \le 0.11$); Qwen2.5-1.5B sits at the bottom of the Qwen family's $\sigmas$ distribution despite the family's overall high-$\sigmas$ tendency.
For each teacher we distill three students {with the body's KD setup} at $\theta = 0.99$ {but a different learning-rate recipe}: budget-matched \pca{}/\uni{}/\inv{} allocations summing to $\sum_l \dthetal$, Minitron-style importance pruning at init, pure KL distillation on SlimPajama at sequence length $2{,}048$, effective batch $256$ sequences, peak LR $3\!\cdot\!10^{-4}$ {with warmup $300$ and a constant learning rate thereafter}, 3K steps.
We report the 5-task average (MMLU, ARC-C, HellaSwag, WinoGrande, PIQA) at step 3K with the same headline-number convention as the body.

Table~\ref{tab:1b-class} reports the results.
{\pca{} scores highest on all three teachers, but every gap is under $+0.6$\pp{}, and on Qwen2.5-1.5B \pca{} leads \inv{} by only $0.04$\pp{} while \inv{} edges \uni{}.}
\begingroup
We do not read these as positive effects.
\endgroup

\begin{table}[t]
\centering
\small
\setlength{\tabcolsep}{4pt}
\caption{\textbf{1B-class 3-way KD results at step 3K.}
5-task averages (MMLU, ARC-C, HellaSwag, WinoGrande, PIQA).
{\pca{} scores highest on every teacher, but every gap is under $+0.6$\pp{} and \inv{} edges \uni{} on Qwen2.5-1.5B; we do not observe a meaningful effect at 1B scale.}}
\label{tab:1b-class}
\begin{tabular}{lccccc}
\toprule
Teacher ($\sigmas$) & \pca{} avg & \uni{} avg & \inv{} avg & $\Delta$ vs.\ \uni & $\Delta$ vs.\ \inv \\
\midrule
Qwen2.5-1.5B (0.105) & $\mathbf{48.25}$ & $48.12$ & $48.21$ & $+0.13$ & $+0.04$ \\
LLaMA-3.2-1B (0.079) & $\mathbf{43.10}$ & $42.88$ & $42.72$ & $+0.22$ & $+0.38$ \\
Gemma-3-1B (0.079)   & $\mathbf{43.50}$ & $43.06$ & $42.97$ & $+0.44$ & $+0.53$ \\
\bottomrule
\end{tabular}

\end{table}

\begingroup
\section{Explained-Variance versus Cosine Readout}
\label{app:ev-readout}

\paragraph{Construction.}
Both readouts use the same centered PCA of $h_l$ (\S\ref{sec:method}).
The explained-variance allocation follows the cumulative explained-variance rule~\citep{jolliffe2002pca}: it sets $w_l$ to the smallest rank at which the cumulative explained variance of layer $l$ reaches $\tau$, with $\tau$ chosen so that $\sum_l w_l$ equals the \pca{} budget $B$ ($\tau^\star = 0.983$ on Qwen2.5-3B and $0.984$ on Gemma-3-4B), rounded to multiples of $128$ with a floor of $128$.
At the nominal $\tau = 0.99$ the explained-variance rule asks for $11$--$12\%$ more total width than the cosine rule on Qwen2.5-3B and Gemma-3-4B.

\paragraph{Where the readouts differ.}
On Qwen2.5-3B and Gemma-3-4B the largest relative differences between the two readouts fall on the layers whose covariance is dominated by a single direction (one on Qwen2.5-3B, two on Gemma-3-4B; Figure~\ref{fig:ev-widths}), where the explained-variance rule cuts width to $2$--$27\%$ of the cosine width; elsewhere the two allocations differ by smaller factors ($0.44$--$1.37\times$), for example on layers~30 and~35 of Qwen2.5-3B and layers~31--33 of Gemma-3-4B.
On a dominant-direction layer the explained-variance rule saturates early, while reconstructing token directions to cosine $0.99$ still requires the remaining components; on Qwen2.5-3B layer~2, for instance, the explained-variance rule is satisfied at one dimension while the cosine at $10\%$ of the width is only $0.73$.
The location of these layers differs between the two teachers: early on Qwen2.5-3B and at mid-depth on Gemma-3-4B.

\begin{figure}[h]
  \centering
  \includegraphics[width=\linewidth]{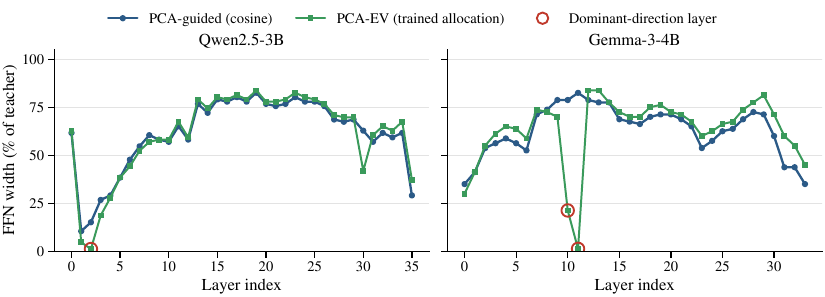}
  \caption{\textbf{Per-layer widths of the cosine (\pca{}) and explained-variance (\ev{}) readouts on Qwen2.5-3B and Gemma-3-4B.}
  Both panels show the trained allocations, budget-matched on each teacher.
  {Open red circles} mark the layers whose covariance is dominated by a single direction (top-1 eigenvalue share above $0.9$): layer~2 of Qwen2.5-3B and layers~10 and~11 of Gemma-3-4B.
  Per-layer correlation of the two width vectors: $r = 0.975$ (Qwen2.5-3B) and $0.31$ (Gemma-3-4B).}
  \label{fig:ev-widths}
\end{figure}

\paragraph{Effect under KD.}
Under the recipes of the body, \ev{} is comparable to \pca{} on Qwen2.5-3B but falls below it on Gemma-3-4B.
On Qwen2.5-3B, \ev{} reaches $60.18 \pm 0.10$ against $60.10 \pm 0.22$ for \pca{} at step 3K, a difference of means of $-0.08$\pp{} (\S\ref{subsec:five-way}).
On Gemma-3-4B, \ev{} reaches $52.60$ at step 3K against $54.24$ for \pca{} and $53.36$ for \uni{}, and stays below both at all eight checkpoints from 1K to 8K (mean $-1.60$\pp{} vs.\ \pca{}), with the largest loss on MMLU ($-3.83$\pp{} at 3K).
Sharply reducing the width of a dominant-direction layer thus has no measurable cost when that layer is layer~2 of Qwen2.5-3B but hurts when the layers sit at mid-depth of Gemma-3-4B{, so layer position appears to matter}.

\paragraph{Correlation with the measured profile.}
The per-layer Pearson $r$ between width vectors (Figure~\ref{fig:ev-widths}) is a coarse summary: it is dominated by the few layers whose widths differ most, and it is undefined for \uni{}.
On Qwen2.5-3B, \slim{} ($r = -0.37$ with \pca{}) scores below both readouts, but on Gemma-3-4B it ($r = -0.61$) scores slightly above \ev{} ($r = 0.31$), and both fall below \uni{}; \ev{} cuts the two dominant-direction layers at mid-depth (layers~10 and~11) to $2{,}176$ and $128$ of $D_l = 10{,}240$ units, against $8{,}064$ and $8{,}448$ for \pca{}.
A single correlation coefficient does not capture such layer-specific cuts.
The cosine readout, which keeps width on these layers, therefore appears to be the more robust allocation rule across teachers.

\endgroup

\begingroup
\section{Serving Cost of a Non-Uniform Student}
\label{app:serving}

A non-uniform width vector is only useful if it can be served.
We implemented the adaptive-width student as a custom vLLM~\citep{kwon2023vllm} model and measured output throughput on the trained Qwen2.5-3B checkpoints, on one V100-SXM2-32GB with tensor parallelism $1$ and CUDA-graph decode: teacher $499.16$ tok/s, \pca{} student $592.83$ ($+18.8\%$ over the teacher), \uni{} student $626.30$ ($+25.5\%$).
No custom kernel is required: the layers remain dense GEMMs that differ in one dimension.
The quality gain therefore carries an explicit serving cost: \pca{} is $5.3\%$ slower than \uni{} at a $0.4\%$ parameter difference ($2.143$B vs.\ $2.152$B); we do not read this as a floor, since the integration performs no per-shape kernel tuning.
\endgroup

\begingroup
\section{Cross-Model Transfer Control}
\label{app:controls}

To separate the measured profile from middle-heaviness as such, we interpolated the profile measured on Qwen3-8B onto the $32$ layers of LLaMA-3-8B and scaled it to the \pca{} budget of LLaMA-3-8B ($335{,}616$); the transferred vector is more strongly middle-heavy than LLaMA-3-8B's own profile and correlates with it at $r = 0.53$ (Figure~\ref{fig:transfer-widths}).
Trained with the LLaMA-3-8B recipe of \S\ref{subsec:kd-setup} alongside the first of the three training seeds in Table~\ref{tab:3way-headline}, it reaches $60.53$ at step 3K, against $61.72$ for \pca{}, $60.85$ for \uni{} and $60.48$ for \inv{}: a more middle-heavy allocation borrowed from another model lands below \uni{} and level with the inverted control.

\begin{figure}[t]
  \centering
  \includegraphics[width=0.7\linewidth]{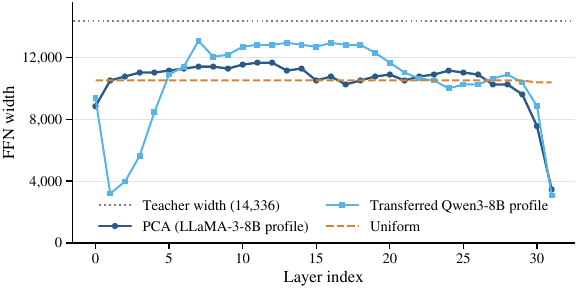}
  \caption{\textbf{Per-layer FFN widths of the LLaMA-3-8B students in the transfer control.}
  All three vectors sum to the same budget ($335{,}616$); the transferred Qwen3-8B profile cuts layers~1--3 to $3{,}200$--$5{,}632$ units, where LLaMA-3-8B's own profile gives $10{,}496$--$11{,}008$.}
  \label{fig:transfer-widths}
\end{figure}
\endgroup

\section{Evaluation-Seed Robustness on LLaMA-3-8B at Step 8K}
\label{app:multi-seed}

To verify that the 3-way KD ordering on LLaMA-3-8B is stable under evaluation-time stochasticity (e.g., few-shot sampling, batching, log-likelihood shuffling) at the longer $8{,}000$-step horizon, we re-evaluate the same step-$8$K \pca{}/\uni{}/\inv{} weights under multiple independent lm-eval-harness seeds.
Each allocation's weights come from a single KD training run; the multi-seed entries {in Table~\ref{tab:multi-seed-llama}} differ only in evaluation seed, not in training.
Table~\ref{tab:multi-seed-llama} reports the resulting 5-task average mean and standard deviation.
The \pca{} $>$ \uni{} $>$ \inv{} ordering holds with evaluation-seed standard deviations under $0.05$\pp{} on the 5-task average, an order of magnitude below the inter-allocation gaps.
The longer $8{,}000$-step horizon {leaves $\Delta(\pca-\uni)$ essentially unchanged relative to step 3K of the same run ($+0.87 \to +0.80$\pp{}) and widens $\Delta(\pca-\inv)$ ($+1.24 \to +1.68$\pp{})}, both remaining {below} the high-$\sigmas$ teachers' gaps.

\begin{table}[t]
\centering
\small
\setlength{\tabcolsep}{5pt}
\caption{\textbf{LLaMA-3-8B 3-way KD at step 8K, multi-seed evaluation.}
$n$ denotes the number of independent evaluation seeds applied to the same single training-run weights for each allocation.
Evaluation-seed standard deviation on the 5-task average is under $0.05$\pp{} for every allocation, an order of magnitude below the inter-allocation gaps ($\Delta(\pca-\uni) = +0.80$\pp, $\Delta(\pca-\inv) = +1.68$\pp).
{Per-task means and standard deviations are reported for MMLU, ARC-C and WinoGrande; HellaSwag and PIQA vary by at most $0.03$ (std).}
This isolates evaluation-time variance only; {training-seed variance at step 3K is reported in \S\ref{subsec:kd-predictors}}.}
\label{tab:multi-seed-llama}
\begin{tabular}{lccccc}
\toprule
Config & \multicolumn{2}{c}{5-task avg} & MMLU & ARC-C & WinoGrande \\
       & mean & std & mean $\pm$ std & mean $\pm$ std & mean $\pm$ std \\
\midrule
\pca{} (n$=$3)      & $\mathbf{64.22}$ & $0.00$ & $\mathbf{50.24} \pm 0.00$ & $\mathbf{45.82} \pm 0.00$ & $\mathbf{70.24} \pm 0.00$ \\
\uni{} (n$=$3)      & $63.42$          & $0.04$ & $48.13 \pm {0.04}$         & $44.97 \pm 0.00$          & $69.74 \pm {0.15}$          \\
\inv{} (n$=$3)      & $62.54$          & $0.03$ & $47.86 \pm {0.12}$         & $43.43 \pm 0.00$          & $67.80 \pm 0.00$          \\
\bottomrule
\end{tabular}

\end{table}

\clearpage

\end{document}